\documentclass[conference]{IEEEtran}
\IEEEoverridecommandlockouts
\usepackage{cite}
\usepackage{amsmath,amssymb,amsfonts}
\usepackage{hyperref}
\usepackage{graphicx}
\graphicspath{ {./images/} }
\usepackage{textcomp}
\usepackage{xcolor}
\usepackage{algorithm}
\usepackage{algpseudocode}
\usepackage{subcaption}
\usepackage{makecell}
\def\BibTeX{{\rm B\kern-.05em{\sc i\kern-.025em b}\kern-.08em
    T\kern-.1667em\lower.7ex\hbox{E}\kern-.125emX}}
\begin{document}

\title{StyleRec: A Benchmark Dataset for Prompt Recovery in Writing Style Transformation}


\author{
    \IEEEauthorblockN{Shenyang Liu, Yang Gao, Shaoyan Zhai, Liqiang Wang}
    \IEEEauthorblockA{
        \textit{Department of Computer Science, University of Central Florida} \\
        Email: shenyang.liu@ucf.edu, yang.gao@ucf.edu, shaoyan.zhai@ucf.edu, liqiang.wang@ucf.edu
    }
}
\IEEEoverridecommandlockouts
\IEEEpubid{\makebox[\columnwidth]{979-8-3503-6248-0/24/\$31.00~\copyright2024 IEEE \hfill} \hspace{\columnsep}\makebox[\columnwidth]{ }}

\maketitle

\IEEEpubidadjcol
\begin{abstract}
Prompt Recovery, reconstructing prompts from the outputs of large language models (LLMs), has grown in importance as LLMs become ubiquitous. Most users access LLMs through APIs without internal model weights, relying only on outputs and logits, which complicates recovery. This paper explores a unique prompt recovery task focused on reconstructing prompts for style transfer and rephrasing, rather than typical question-answering. We introduce a dataset created with LLM assistance, ensuring quality through multiple techniques, and test methods like zero-shot, few-shot, jailbreak, chain-of-thought, fine-tuning, and a novel canonical-prompt fallback for poor-performing cases. Our results show that one-shot and fine-tuning yield the best outcomes, but highlight flaws in traditional sentence similarity metrics for evaluating prompt recovery. Contributions include (1) a benchmark dataset, (2) comprehensive experiments on prompt recovery strategies, and (3) identification of limitations in current evaluation metrics, all of which advance general prompt recovery research, where the structure of the input prompt is unrestricted.
\end{abstract}

\begin{IEEEkeywords}
Prompt Recovery, Language Model Inversion, LLM, Adversarial Attack
\end{IEEEkeywords}

\section{Introduction}
Large Language Models (LLMs) have become essential to various applications due to their ability to generate high-quality outputs based on user prompts. However, there are instances where we only have access to the generated output but need to identify the corresponding prompt to get that specific output. This task, known as ``Prompt Recovery," was introduced by \cite{morris2023language} in the context of closed-source LLMs. Subsequent studies have addressed this challenge as a form of attack, such as prompt leakage or jailbreak attempts \cite{sha2024prompt} \cite{wu2023jailbreaking}, highlighting the security implications of recovering prompts to defend against malicious uses of LLMs. Successful prompt recovery is crucial for mitigating risks associated with harmful prompt generation \cite{chao2023jailbreaking}, determining user liability \cite{skapars2024slander}, and verifying potential copyright violations \cite{karamolegkou2023copyright}.

The fundamental difficulty of prompt recovery lies in the fact that exact inversion of outputs to prompts typically requires additional information, like the full probability distribution, which is only available for some LLMs \cite{morris2023language}. For models that are accessible only through inference APIs, the information is restricted. Additionally, in scenarios where outputs are derived from documents without access to the original prompt or supplementary data, the challenge of prompt recovery becomes even more evident.

While most existing work in this area focused on question-answering datasets \cite{morris2023language} \cite{give2024uncovering} \cite{gao2024dory}, our research explores a specialized scenario in which prompts are used to transform writing styles or rephrase sentences. The task involves recovering the transformation prompt from the original sentence and its corresponding output. Different from \cite{chen2024advancing}, our work focuses on providing an open-source dataset along with a detailed methodology for its construction and testing method within a single model for this task.

In this paper, we introduce a benchmark dataset, named \textbf{StyleRec}\footnote{For the dataset, please refer to the following GitHub repository:
\href{https://github.com/promptrecovery501/StyleRec}{https://github.com/promptrecovery501/StyleRec}.}, which ensures quality and diversity through rigorous construction techniques. We detail the dataset's creation process to facilitate further research. Additionally, we evaluate five different methods to determine the most effective approach for prompt recovery in this specialized context.
Our contributions are as follows. (1) We present the first benchmark dataset with detailed construction guidelines, enabling researchers to generate additional data. (2) Our experimental results demonstrate the effectiveness of specific methods, offering guidance for future research in this domain. (3) We identify flaws in commonly used sentence similarity metrics when applied to the prompt recovery task. Additionally, we highlight the unique challenges of prompt recovery in different scenarios, underscoring the complexity of the general prompt recovery task where the format of the prompt is unrestricted.

\begin{figure*}[h]
\centering
\includegraphics[width=\textwidth]{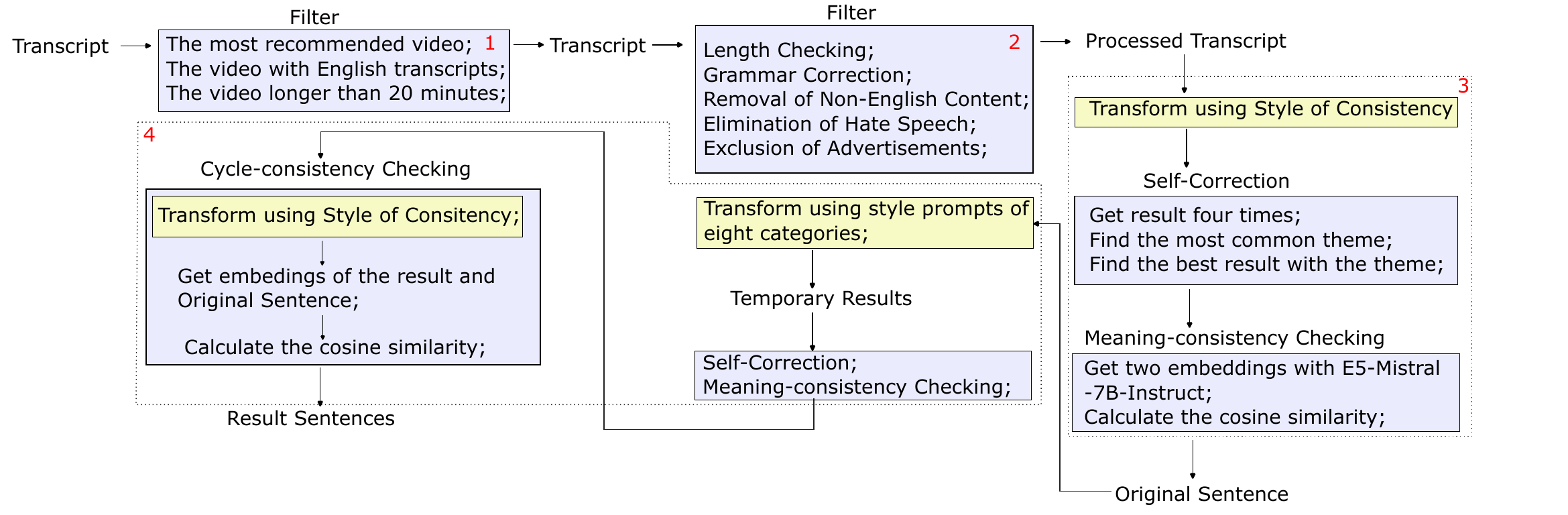}
\caption{The Workflow of Data Generation}
\label{fig:data_generation}
\end{figure*}

\section{Related Work}
\subsection{Language Model Jailbreaking}
Language Model Jailbreaking refers to techniques used to bypass or undermine the safety and ethical guidelines embedded within LLMs. \cite{wu2023jailbreaking} employs a Self-Adversarial Attack and demonstrates that paraphrasing a system prompt can effectively bypass a target model’s safeguards. \cite{liu2023jailbreaking} conducts an empirical study that categorizes jailbreak methods into three primary strategies: Pretending, Attention Shifting, and Privilege Escalation. The work in \cite{wei2024jailbroken} identifies two primary causes for LLM safeguard failures: competing objectives and generalization mismatch. These factors have inspired the development of various jailbreak techniques \cite{deng2023jailbreaker} \cite{mazeika2024harmbench}.

However, most evaluations of jailbreak methods have not thoroughly considered defending LLMs. The study by \cite{xu2024comprehensive} addresses this gap by assessing the effectiveness of both jailbreak attacks and defense techniques. The findings across these papers underscore a critical point: no single defense can block all attacks, and no attack can penetrate all defenses. Nevertheless, the ongoing battle between attackers and defenders drives the continuous improvement of LLMs.

\subsection{Model Stealing}
\cite{tramer2016stealing} first formalizes the Model Stealing problem that is to steal the LLM’s weights through interaction with the LLM itself. The approach has been used in different tasks ,such as membership inference \cite{shokri2017membership}, federate learning \cite{kairouz2021advances} and machine translation \cite{wallace2020imitation} and different areas, such as healthcare\cite{miotto2018deep}, biology\cite{ching2018opportunities} and cryptography\cite{mishra2020delphi}. Recent studies \cite{gudibande2023false} \cite{morris2023language} suggest that reconstructing model weights may replicate models capability of imitating surface syntax, but have difficulty to restore their underlying decision-making mechanisms. 

Since many model stealing methods are developed, protection methods are also improved by many work. \cite{juuti2019prada}\cite{lee2019defending} focused on the defense of normal deep neurual network. \cite{sebastian2023privacy} discussed strategies of protection for chatbots. 

Beyond the protection methods that make the model stronger, another method to avoid issues is to detect model stealing \cite{guan2022you} \cite{liu2024model} before submitting the inputs into the model.

\subsection{Prompt Recovery}
The term ``Prompt Recovery" has been recently introduced in studies such as \cite{gao2024dory}, \cite{give2024uncovering}, and \cite{chen2024advancing}. However, the underlying concept has been explored for many years under the name ``Model Inversion" in the computer vision domain \cite{fredrikson2015model} \cite{zhang2020secret}. \cite{morris2023language} applied model inversion techniques to LLMs, highlighting the crucial role of logits (e.g., probability distributions) in successful prompt recovery.

\cite{give2024uncovering} focused on creative writing, such as poetry, and employed a method termed ``sample inverter" by \cite{morris2023language}, which does not rely on logits. Building on this, \cite{gao2024dory} delved deeper into the relationship between output probability-based uncertainty and the effectiveness of prompt recovery methods. They developed a novel approach that leverages this uncertainty through a chain-of-thought methodology.

Simultaneously with our work, \cite{chen2024advancing} developed a similar idea, focusing on a specialized scenario where the goal is to predict the prompt that alters the writing style or rephrases an original sentence, given the original sentence and its corresponding output. Previous studies \cite{morris2023language}, \cite{give2024uncovering} and \cite{gao2024dory} have primarily concentrated on using the LLM's output to predict the original prompt or question.

\section{Data Generation with LLM}\label{Data_Generation_with_LLM}
\begin{figure}[h]
\centering
\includegraphics[width=\columnwidth]{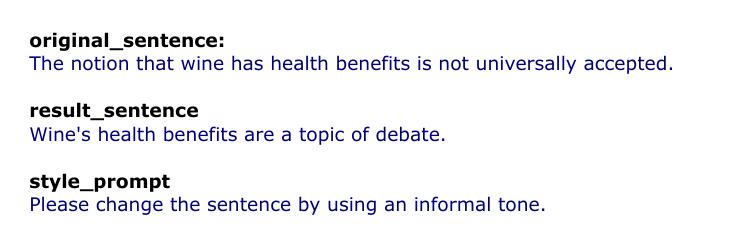}
\caption{Sample Data}
\label{fig:sample_data}
\end{figure}

In this section, we describe the process used to generate our dataset (see Figure \ref{fig:data_generation}). First, we selected YouTube videos covering various topics relevant to daily life, extracted the transcripts from these videos (Section~\ref{Preparation}), and applied multiple filters to ensure data quality (Section~\ref{Preprocessing}). Using the preprocessed transcripts, we crafted prompts to transform the style of these transcripts (Section~\ref{Methods_for_Generation}). The resulting data was further refined using validation methods based on cosine similarity (Section~\ref{Data_Validation}). The sample data is shown in Fig. \ref{fig:sample_data}.

\subsection{Data Preparation}\label{Preparation}
We utilize YouTube video transcripts as the primary source for generating our dataset. The platform provides both automatically generated and manually reviewed transcripts, making it an ideal and comprehensive data source for this study. The inclusion of such a wide variety of content ensures that the dataset is rich and multifaceted, enhancing the generalizability and broad applicability of the research to various real-world use cases.

To ensure that our dataset captures a wide spectrum of everyday scenarios, we selectively extract transcripts from YouTube videos categorized into diverse topics, including: travel, education, entertainment, environment, fashion, finance, food, health, history, law, news, real estate, family, religion, science, culture, sports, and technology. This deliberate categorization ensures that the dataset reflects a range of human experiences, perspectives, and conversational styles. By incorporating content from various categories, we aim to guarantee that the dataset covers both formal and informal language, technical and non-technical topics, and different levels of complexity in conversation. 

To identify suitable videos for each category, we apply the following criteria as filters for selection: (1) the video must be the most recommended within its specific category, ensuring that the content is both relevant and engaging, (2) the video must have English transcripts, ensuring language consistency throughout the dataset, and (3) the video must be longer than 20 minutes, providing enough content to generate substantial transcripts for analysis. These filters ensure that the dataset is not only diverse but also provides enough contextual depth and richness for effective model training and testing.

\subsection{Data Preprocessing}\label{Preprocessing}
\begin{figure}[h]
\centering
\includegraphics[width=\columnwidth]{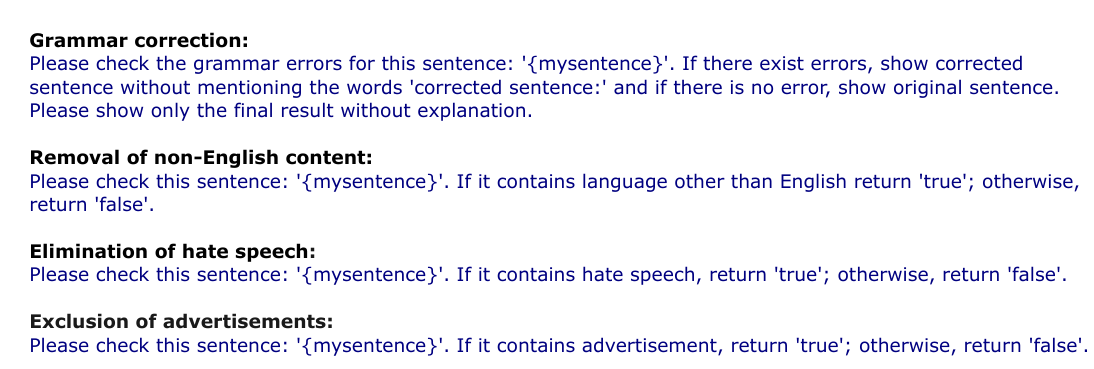}
\caption{Prompts for filters}
\label{fig:prompts_preprocessing}
\end{figure}

After the initial selection of videos, we preprocess the transcripts to prepare them for use in the study. Fig. \ref{fig:data_generation} illustrates the filtering criteria used in this preprocessing phase. First, we remove any special characters and extraneous information, such as the speaker’s name that often precedes dialogue in transcripts. This step is crucial to eliminate non-conversational elements that may interfere with the language modeling process. The transcripts then undergo a series of five additional filtering steps: length checking, grammar correction, non-English content removal, hate speech elimination and advertisement exclusion. Fig. \ref{fig:prompts_preprocessing} shows the specific prompts and rules we used to implement these filtering steps. Once these preprocessing steps are completed, we receive a set of clean, high-quality transcripts that are ready for use in our study.

Finally, these cleansed transcripts are input into the LLM using a variety of style transformation prompts. The LLM generates outputs based on the input transcripts, applying different stylistic changes.

\subsection{Methods for Generation}\label{Methods_for_Generation}
\begin{figure}[h]
\centering
\includegraphics[width=\columnwidth]{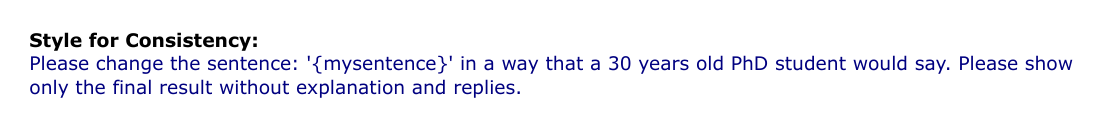}
\caption{Prompts for consistency}
\label{fig:prompts_consistency}
\end{figure}

\begin{figure}[h]
\centering
\includegraphics[width=\columnwidth]{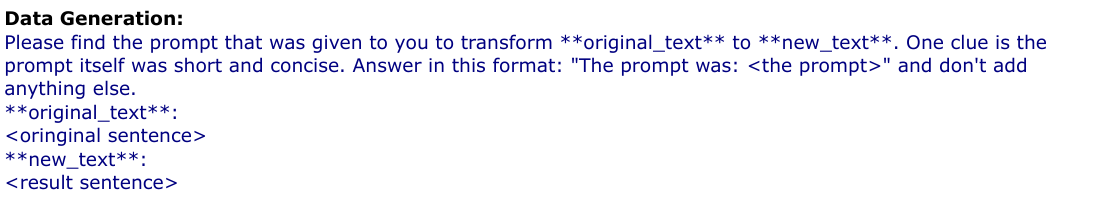}
\caption{Prompts for Data Generation}
\label{fig:prompts_generation}
\end{figure}

\begin{figure}[h]
\centering
\includegraphics[width=\columnwidth]{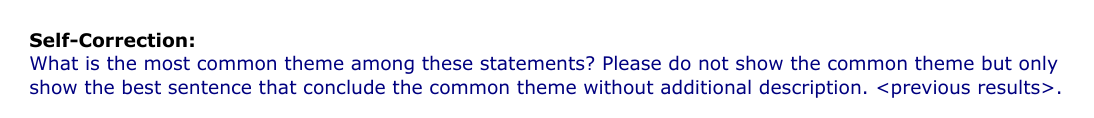}
\caption{Prompts for Data Validation}
\label{fig:prompts_validation}
\end{figure}

With the preprocessed transcripts, we generated our dataset by applying style transformation prompts and obtaining the outputs from the LLM. Each instance in the dataset follows the format: original sentence, result sentence, style prompt. The prompts we employed include eight categories: tone, family roles, occupation, celebrity, historical periods, passive voice, diary style, and proverbs. All 33 styles are shown in Table~\ref{all_styles}. Given the varied sources of the transcripts, the styles were initially inconsistent. The prompt template is shown in Fig. \ref{fig:prompts_generation} We standardized the style of the transcripts to a uniform ``style for consistency" (Fig. \ref{fig:prompts_consistency}), which was used as the original sentence in our dataset. To ensure the stability of the results, multiple outputs were generated for each style prompt. We then applied self-correction \cite{ahmed2023better} to select the best output, the prompt is shown in Fig. \ref{fig:prompts_validation}. For further study, we also collected logits for next-token probability and Length-normalized Predictive Entropy (LN-PE) \cite{malinin2020uncertainty} for use in few-shot sample selection.

\subsection{Data Validation}\label{Data_Validation}
To validate the generated data, we measured the cosine similarity between the original sentence and the output after style transformation, as a metric referred to as meaning consistency. To differentiate the current generated data with the final results, we call it ``temporary results". As described in Sec~\ref{Methods_for_Generation}, we first transformed the transcripts to the ``style for consistency" as the original sentence before generating the result sentence using the style transformation prompt. Subsequently, the result sentence was converted back to the ``style for consistency" to obtain the predicted original sentence. We then measured the cosine similarity between the original sentence and the predicted original sentence, a process known as cycle consistency \cite{chen2020canzsl}. Finally, we applied thresholds for both meaning consistency and cycle consistency to filter out inconsistent results.

\begin{table}[!htb]
    \centering
    \caption{All styles in different categories}\label{all_styles}
    \resizebox{\columnwidth}{!}{%
    \begin{tabular}{ll}
    \hline
    \textbf{Category} & \textbf{Styles} \\
    \hline
    Tone & \makecell[l]{formal, informal, optimistic, pessimistic, humorous, serious, \\ inspiring, authoritative, persuasive} \\
    Family Roles & grandfather, grandmother, father, mother, son, daughter \\
    Occupation & \makecell[l]{professor, doctor, policeman, priest, kindergarten teacher, \\ businessman} \\
    Celebrity & \makecell[l]{Donald Trump, Joe Biden, Ellen DeGeneres, Kevin Hart, \\ Conan O'Brien, Steve Harvey} \\
    Historical Period & old English, middle English, early modern English \\
    Passive & passive \\
    Diary & diary \\
    Proverb & proverb \\
    \hline
    \end{tabular}%
    }
\end{table}

\begin{figure}[h]
\centering
\includegraphics[width=\columnwidth]{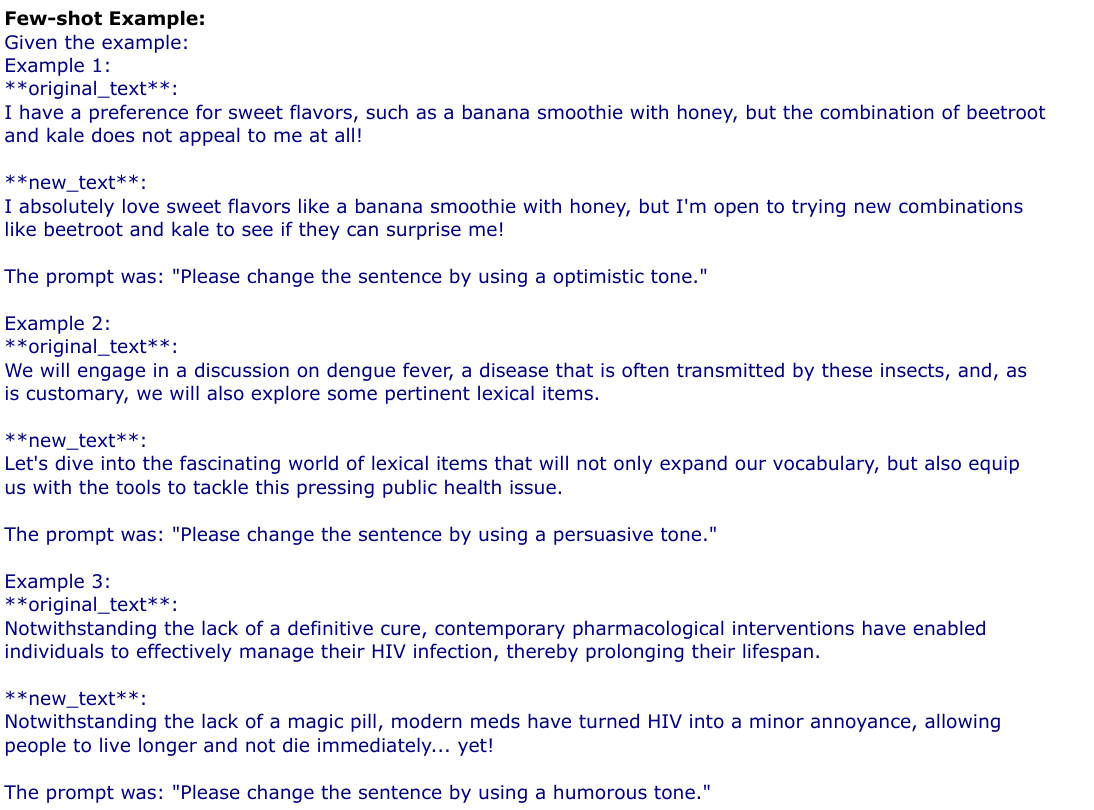}
\caption{Prompts for Few-shot}
\label{fig:prompts_fewshot}
\end{figure}

\section{Methods for Prompt Recovery}
\subsection{Direct Inference with LLM}
In earlier models, limited size and training on specific tasks meant that performance on new tasks without fine-tuning was often poor. However, with the advent of modern LLMs, which are trained on large amounts of data and a wide range of downstream tasks, direct inference has become feasible. For this study, we manually crafted prompts for both zero-shot (see Fig.\ref{fig:prompts_generation}) and few-shot (see Fig.\ref{fig:prompts_fewshot}) settings to generate the result sentences. The size of examples for the few-shot setting is kept constrained because longer sample sizes lead to increased inference time. We selected data with the highest LN-PE scores for each different style as sample data and randomly picked a subset for few-shot learning.

\subsection{Jailbreak}
As mentioned in the results of \cite{gao2024dory}, jailbreak prompt does not help much for prompt recovery task. We do not try all the jailbreak prompts in \cite{gao2024dory}, but only use Prefix Injection and Refusal Suppression discussed in \cite{wei2024jailbroken}.

\subsection{Chain of Thoughts}
Chain of Thoughts (CoT) \cite{wei2022chain} is effective for enhancing reasoning in language models because it breaks down complex problems into smaller, manageable steps, leading to more interpretable and accurate responses. In this work, we break the problem into four steps: compare tone and style, identify the changes, check the purpose of the transformation, and consider the clues to get the final result.

\subsection{LLM Fine-Tuning}
Although LLMs can address the problem through direct inference with some samples, fine-tuning enhances the model's ability to learn from training data. To efficiently fine-tune the model while limiting the training time, we employed Low-Rank Adaptation (LoRA) \cite{hu2021lora}, a parameter-efficient technique for fine-tuning. This method adjusts only a small subset of the model’s trainable parameters while maintaining strong performance.

\subsection{Canonical Prompt for Abnormal Outputs}
The ``Canonical Prompt" concept involves identifying a prompt that is close to the majority of prompts in the training set $S_{T}$, enabling it to serve as a fallback prompt that avoids poor performance across various inputs. 
The approach is summarized in Algorithm~\ref{alg:canonical_prompt}. We detail the process for generating a canonical prompt as follows:
First, we manually create seed prompts and add them to the Generated Prompt Set $S_{G}$. We then calculate the cosine similarity between sampled prompts $S_{Sample}$ from $S_{T}$ and $S_{G}$ and determine the average similarity for each prompt in $S_{G}$. Prompts in $S_{G}$ with high similarity are retained as our new training set, $S_{G_{new}}$, we keep top $k$ of the prompts. Next, we use $S_{G_{new}}$ to generate a Vocabulary Set $S_{V}$ and employ beam search to insert words from $S_{V}$ into the prompts in $S_{G}$, cosine similarity is used as first step to evaluate generated prompts. New prompts that significantly increase similarity to $S_{G_{new}}$ are stored as coarse results. Since word insertion can result in excessively long prompts, we apply a threshold to trim them. To further reduce the search space, we use a greedy search approach, inserting words only into the current best prompt. To improve the performance, all the steps discussed above can be run multiple times before get the best prompt. Finally, we get the final results and find the best prompt to return.

\begin{algorithm*}
\caption{Canonical Prompt}\label{alg:canonical_prompt}
\hspace*{\algorithmicindent} \textbf{Input} $ S_{T}, SeedPrompts, LoopTimes$ 
\begin{algorithmic}
\State $S_{G} \gets SeedPrompts$
\State $i=0$
\While{\texttt{$i < LoopTimes$}}
\For{\texttt{$sentence_g \gets$ enumerate($S_{G}$)}}
\State $Similarity_g \gets 0$
\State $S_{Similarity} \gets \emptyset$ \Comment{$S_{Similarity}$ is the set for average similarity between one prompt in $S_{G}$ and each prompt in $S_{Sample}$}
\State $S_{Sample} \gets$ \textbf{Sample}$(S_{T})$
\For{\texttt{$sentence_{s} \gets$ enumerate($S_{Sample}$)}}
\State $Similarity_g \gets Similarity_g +   
 $CosineSimilarity$(sentence_g, sentence_{s})$
\EndFor
\State $Similarity_g \gets$ Average$(Similarity_g)$
\State $S_{Similarity} \gets S_{Similarity} \cup Similarity_g$
\EndFor
\State $S_{G_{new}} \gets$ \textbf{GetTopk}$(S_{Similarity})$ \Comment{Get k prompts with the best similarity}
\State $S_{V} \gets$ \textbf{GenerateVocabulary}$(S_{G_{new}})$
\For{\texttt{$sentence_g \gets$ enumerate($S_{G_{new}}$)}}
\State $CoarseResults \gets $\textbf{BeamSearch}$(sentence_g, S_{V})$
\State $Results \gets$ \textbf{Trim}$(CoarseResults)$
\EndFor
\State $i=i+1$
\State $ S_{G} \gets S_{G} \cup Results$
\EndWhile
\State $BestPrompt \gets$ \textbf{GetBest}$(Results)$
\State \Return $BestPrompt$
\end{algorithmic}
\end{algorithm*}

\section{Experimental Setup}
\subsection{LLM}
For inference and fine-tuning, we conducted experiments using Llama 3 8B model \cite{dubey2024llama} and Mistral 7B v0.3 model \cite{jiang2023mistral}. We train models for 1 epoch with Paged Adam 8Bit optimizer with a learning rate of $2e - 4$. To generate embedding vectors for cosine similarity, we utilize the E5-Mistral-7B-Instruct model \cite{wang2023improving}, which is recognized as one of the top-performing embedding models. We use a constant learning rate with linear warmup over the first 30\% training steps. We train in FP32 precision.

\subsection{Dataset}
We generated the dataset as described in Section~\ref{Data_Generation_with_LLM}. First we get 16174 transcripts from selected YouTube videos, and use filters to get 13686 filtered transcripts. We did not create results for each of the filtered transcript with all 33 transformation styles, instead, we just use one of these styles for them. Then, we get the dataset consisting of 13686 instances across various categories, each including the original sentence, result sentence, style prompt, logits, and LN-PE values. We only take the data with both meaning consistency and cycle consistency larger than 0.75, so we finally get 10193 instances. The dataset is then divided as follows: 80\% for training, 10\% for validation, and 10\% for testing.

\subsection{Evaluation Metrics}
We use sharpened cosine similarity (SCS) as in \cite{chen2024advancing} and we also experiment with Exact Match, BLEU-4\cite{papineni2002bleu}, Rouge-L\cite{lin2004rouge} and F1 score at the token level as in \cite{morris2023language} and \cite{gao2024dory}.
\subsubsection{Rouge-L}
ROUGE-L is a metric used to evaluate the quality of summaries by comparing them to reference summaries. Specifically, it utilizes the Longest Common Subsequence (LCS) between the prediction (pred) and ground truth sentence (gt).
\begin{equation}\label{Rouge-L-1}
P=\frac{LCS(pred, gt)}{length(pred)},
\end{equation} where $LCS(pred,gt)$ is the length of the longest common subsequence between prediction and ground truth sentence, $length(pred)$ is the total number of words in the prediction.
\begin{equation}\label{Rouge-L-1}
R=\frac{LCS(pred, gt)}{length(gt)},
\end{equation} where $length(gt)$ is the total number of words in the ground truth sentence.
\begin{equation}\label{Rouge-L-1}
F1=\frac{2 \times P \times R}{P + R}
\end{equation}
We use ROUGE-L F1 score in our result.

\subsubsection{Token F1}
Token F1 is a metric used in tasks where predictions are made at the token level. The result is calculate by comparing predictions and ground truth sentences.
\begin{equation}
P=\frac{TP}{TP + FP},
\end{equation} where $TP$ stands for the number of words shared by prediction and ground truth sentence and $FP$ stands for the number of words in the ground truth sentence but not in the prediction.

\begin{equation}
P=\frac{TP}{TP + FN},
\end{equation} where $FN$ stands for the number of words in the prediction but not in the ground truth sentence.

\begin{equation}\label{Rouge-L-1}
F1=\frac{2 \times P \times R}{P + R}
\end{equation}

\subsubsection{Sharpened Cosine Similarity}
Unlike the work of \cite{morris2023language} and \cite{gao2024dory} that just use cosine similarity, we employed sharpened cosine similarity (SCS) to provide a more refined similarity score as in \cite{chen2024advancing}. The similarity is calculated as follows: 
\begin{equation}\label{SCS}
SCS(v_{original}, v_{result})=(\frac{v_{original} \cdot v_{result}}{\Vert v_{original} \Vert \Vert v_{result} \Vert})^3, 
\end{equation} where $v_{original}$ and $v_{result}$ are the embedding vectors of the original sentence and the results, respectively. These vectors are generated using the E5-Mistral-7B-Instruct model \cite{wang2023improving}. We opted not to use the Sentence-T5-Base model as in \cite{chen2024advancing} for two main reasons. First, the E5-Mistral-7B-Instruct model outperforms Sentence-T5-Base. Secondly, the Sentence-T5-Base model implemented by Hugging Face has a known issue with the embedding of the word ``lucrarea," which is almost identical to the embedding of the special token ``\textless/s\textgreater," used to close output sentences. If ``lucrarea" is added multiple times to the input, the output will contain many ``\textless/s\textgreater" tokens. When calculating the similarity between such outputs and the ground truth sentences, the similarity is likely to be inflated compared to outputs generated without ``lucrarea" due to the presence of ``\textless/s\textgreater" in all the embeddings.
\subsubsection{BLEU}
BLEU is an evaluation metric used primarily for assessing the quality of machine-translated text by comparing it to one or more ground truth translations. It measures the correspondence between the machine-generated output and the ground truth translations using n-gram precision (e.g., contiguous sequences of words of length n). The BLEU is calculate by the equation as follows:
\begin{equation}
\text{BLEU} = BP \times \exp \left( \frac{1}{N} \sum_{n=1}^{N} \log p_n \right),
\end{equation}
where $BP$ is the brevity penalty, $N$ is the maximum n-gram length and $p_n$ is the modified precision for n-grams of length $n$.
And the brevity is calculated as follows:
\begin{equation}
BP = \begin{cases}
1 & \text{if } pred > gt \\
e^{1 - \frac{gt}{pred}} & \text{if } pred \leq gt
\end{cases},
\end{equation}
where pred is the length of the prediction, gt is the length of the ground truth sentence.

\subsubsection{Exact Match}
Exact Match (EM) is a simple and intuitive evaluation metric that measures how often a model's prediction exactly matches the ground truth.

\section{Experimental Results and Analysis}
\begin{table}[!htb]
    \centering
    \caption{Experimental Results with Mistal-7B-Instruct Model}\label{result1}
    \begin{tabular}{llll}
    \hline
    \textbf{Setting} & \textbf{Rouge-L} & \textbf{Token F1} & \textbf{SCS}\\
    zero-shot & 14.25 & 14.27 & 76.73\\
    jailbreak+zero-shot & 15.00 & 15.01 & 75.7\\
    CoT+zero-shot & 15.34 & 13.94 & 76.68\\
    canonical prompt+zero-shot & 14.34 & 14.36 & 76.78\\
    one-shot & 32.06 & 31.88 & 84.88\\
    jailbreak+one-shot & 26.40 & 26.01 & 81.40\\
    CoT+one-shot & 31.24 & 30.34 & \textbf{84.98}\\
    canonical prompt+one-shot & 32.08 & 31.89 & 84.90\\
    three-shot & 19.98 & 19.65 & 73.89\\
    jailbreak+three-shot & 16.90 & 16.18 & 76.60\\
    CoT+three-shot & 30.36 & 29.42 & 81.42\\
    canonical prompt+three-shot & 20.50 & 20.10 & 74.34\\
    five-shot & 21.40 & 21.16 & 74.91\\
    jailbreak+five-shot & 16.25 & 15.58 & 76.53\\
    CoT+five-shot & 31.51 & 30.41 & 81.66\\
    canonical prompt+five-shot & 21.97 & 21.64 & 75.32\\
    fine-tuning & \textbf{32.80} & \textbf{36.39} & 84.58\\
    \hline
    \end{tabular}
\end{table}

\begin{table}[!htb]
    \centering
    \caption{Experimental Results with Meta-Llama-3-8B-Instruct Model}\label{result2}
    \begin{tabular}{llll}
    \hline
    \textbf{Setting} & \textbf{Rouge-L} & \textbf{Token F1} & \textbf{SCS}\\
    zero-shot & 15.34 & 14.88 & 81.80\\
    jailbreak+zero-shot & 16.98 & 16.45 & 80.51\\
    CoT+zero-shot & 12.00 & 10.63 & 77.98\\
    canonical prompt+zero-shot & 15.34 & 14.88 & 81.80\\
    one-shot & \textbf{79.66} & \textbf{79.64} & \textbf{90.56}\\
    jailbreak+one-shot & 36.97 & 36.60 & 86.88\\
    CoT+one-shot & 42.15 & 41.26 & 84.68\\
    canonical prompt+one-shot & 79.66 & 79.64 & 90.56\\
    three-shot & 68.42 & 68.26 & 90.20\\
    jailbreak+three-shot & 37.77 & 36.98 & 87.27\\
    CoT+three-shot & 12.51 & 10.88 & 79.02\\
    canonical prompt+three-shot & 68.42 & 68.26 & 90.20\\
    five-shot & 57.46 & 56.60 & 87.83\\
    jailbreak+five-shot & 35.42 & 35.24 & 85.37\\
    CoT+five-shot & 10.74 & 9.34 & 78.16\\
    canonical prompt+five-shot & 57.46 & 56.60 & 87.83\\
    fine-tuning & 17.18 & 16.88 & 82.12\\
    \hline
    \end{tabular}
\end{table}

\subsection{Overall Result and Analysis}
Since the BLEU and Exact Match scores are either zero or very close to zero, we choose not to include them in the result tables. The results are shown in Table~\ref{result1} and Table~\ref{result2}. Focusing on the performance of two different models, the one-shot setting and fine-tuning deliver the best results. The one-shot setting achieves the highest SCS scores for both models, while fine-tuning yields the best Rouge-L and Token F1 scores for Mistal-7B-Instruct. Overall, Meta-Llama-3-8B-Instruct outperforms Mistal-7B-Instruct in most settings, except for CoT in the zero-shot, three-shot, and five-shot settings, where Mistal-7B-Instruct has an average of 13.99 higher for Rouge-L, 14.31 higher for Token F1, and 1.3 higher for SCS compared to Meta-Llama-3-8B-Instruct.

It is clear that Meta-Llama-3-8B-Instruct is the best model for our task, while the CoT technique is proven less beneficial for it. This suggests that Meta-Llama-3-8B-Instruct is able to derive better solutions for the task independently, without relying on the ``thoughts" manually designed for CoT.

Next, we examine the performance of each method for the two models, using the zero-shot setting as a baseline for comparison.

First, although all few-shot settings outperform zero-shot, the one-shot setting leads to the most significant improvements. For Mistal-7B-Instruct, one-shot improves Rouge-L by 17.91, Token F1 by 17.61, and SCS by 12.15. For Meta-Llama-3-8B-Instruct, one-shot boosts Rouge-L by 64.32, Token F1 by 64.76, and SCS by 8.76.

Secondly, the jailbreak setting results in a performance drop, with Mistal-7B-Instruct showing an average decrease of 3.29 for Rouge-L, 3.55 for Token F1, and 0.07 for SCS. Similarly, Meta-Llama-3-8B-Instruct experiences an average decrease of 23.44 for Rouge-L, 23.53 for Token F1, and 1.11 for SCS.

Thirdly, CoT improves Mistal-7B-Instruct’s performance with an average increase of 5.19 points for Rouge-L, 4.29 points for Token F1, and 3.58 points for SCS. In contrast, Meta-Llama-3-8B-Instruct's performance shows a decrease of 35.87 for Rouge-L, 36.82 for Token F1, and 7.64 for SCS.

Fourthly, the canonical prompt setting provides only slight improvements for Mistal-7B-Instruct, with average gains of 0.3 for Rouge-L, 0.26 for Token F1, and 0.23 for SCS. For Meta-Llama-3-8B-Instruct, there is no noticeable change.

Finally, fine-tuning offers significant improvements for Mistal-7B-Instruct, with gains of 18.55 for Rouge-L, 22.12 for Token F1, and 7.85 for SCS. For Meta-Llama-3-8B-Instruct, fine-tuning provides more modest improvements: 1.84 points for Rouge-L, 2.00 points for Token F1, and 0.42 points for SCS. Surprisingly, fine-tuning Meta-Llama-3-8B-Instruct does not surpass the one-shot setting as seen in the few-shot settings as well. We dicuss the reason in the next section.

In conclusion, the one-shot setting delivers strong results. Methods like jailbreak, canonical prompt, and CoT offer slight improvements for Mistal-7B-Instruct but have little to no impact on Meta-Llama-3-8B-Instruct. Ultimately, Meta-Llama-3-8B-Instruct with one-shot setting proves to be the superior setting for our prompt recovery task in our study.

\subsection{Error Analysis}

\begin{table*}[!htb]
    \centering
    \caption{Samples of Groud Truth for Error Analysis}\label{ground_truth}
    \begin{tabular}{ll}
    \hline
    \textbf{scenario} & \textbf{Ground Truth}\\
    low score with acceptable answer & Please change the sentence by using a businessman's style.\\
    high score with incorrect answer & Please change the sentence by using a mother's style.\\
    low score with incorrect answer & Please change the sentence by using a doctor's style.\\
    \hline
    \end{tabular}
\end{table*}

\begin{table*}[!htb]
    \centering
    \caption{Samples of Prediction and Metrics for Error Analysis}\label{prediction}
    \begin{tabular}{llll}
    \hline
    \textbf{Prediction} & \textbf{Rouge-L} & \textbf{Token F1} & \textbf{SCS}\\
    Rewrite the text to make it more concise and formal. & 10.53 & 10.53 & 85.17\\
    Please change the sentence by using a son's style. & 88.89 & 88.89 & 95.25\\
    Rewrite the text to focus on the concept of cultural exchange and innovation. & 9.52 & 9.09 & 70.43\\
    \hline
    \end{tabular}
\end{table*}

Given the variety of experimental settings we tested, it is challenging to review all errors comprehensively. Here, we focus on the errors observed in the one-shot and fine-tuning settings. (See Table~\ref{ground_truth} and Table~\ref{prediction})

The first scenario involves low scores with acceptable answers. For example, the ``businessman's style" prompt often results in outputs that are concise and formal, which are acceptable for most people. However, the scores for these outputs are consistently low across all metrics. This type of error is common in the fine-tuning results.

The second scenario involves high scores with incorrect answers. For instance, the model predicts ``son" instead of ``mother" in a family role task. The rest of the sentence is correctly predicted and the overall score remains high, even though the output is clearly wrong from a human perspective. This highlights a limitation in the metrics, as they fail to penalize such errors adequately. This type of error is common in the one-shot setting.

The third scenario involves low scores with incorrect answers, which is expected based on the metrics we use. However, we cannot definitively say that the answer is 100\% wrong, as even though we generate the result based on the ground truth, prompts other than the ground truth may also lead to similar result sentences.

In the first scenario, Rouge-L and Token F1 fail to reflect the close semantic similarities between the prediction and ground truth, while SCS captures some similarity but remains inadequate. In the second scenario, we conclude that Rouge-L, Token F1, and SCS all yield high scores, indicating that they do not fully capture the nuances of our specific task. In the third scenario, it underscores the inherent difficulty of the prompt recovery task, emphasizing the need for further research and exploration.

\section{Limitations}
First, as mentioned in error analysis, the metrics we use have defects when facing some specific scenarios and need to be improved. Secondly, although LLMs demonstrate strong performance on our dataset, this success may not extend to out-of-distribution data, highlighting the need for further experiments to assess generalization. Thirdly, given the vast scale of LLMs, the dataset we generated may be insufficient compared to the data used during pre-training for these LLMs. Expanding the dataset, potentially through data augmentation techniques, could improve performance. Fourthly, our research focuses on English transcripts from YouTube videos, which provide a rich dataset for understanding contemporary language usage in various contexts, but the dataset can be extend to other languages to incorporate multilingual and cross-cultural perspectives. Fifthly, our study focuses on a specific prompt recovery scenario, and many of the methods we explored may not easily apply to more general prompt recovery tasks, where the output format is not constrained. Addressing the general prompt recovery challenge may require leveraging adversarial attack techniques, jailbreak methods, or other strategies to extract additional information about the input prompt.

\section{Conclusion}
In conclusion, we explore a unique aspect of the prompt recovery task, focusing on scenarios where a prompt alters the writing style or rephrases a sentence. Our work introduces a new benchmark dataset, \textbf{StyleRec}, specifically designed to address this specialized challenge, ensuring both quality and comprehensive coverage. Through our experiments, we demonstrate that some methods contribute to this complex problem and one-shot is the best among them. However, the error analysis reveals that the current metrics are inadequate for our specific task and require improvement. Our research not only advances the understanding of prompt recovery but also opens up new avenues for further exploration, encouraging the development of innovative approaches to tackle the general prompt recovery challenge.

\section*{Acknowledgment}
The work was supported in part by NSF under Grants 2321572 and 1952792.
\renewcommand{\thefootnote}{} 
\footnotetext{Citation: S. Liu, Y. Gao, S. Zhai and L. Wang, "StyleRec: A Benchmark Dataset for Prompt Recovery in Writing Style Transformation," 2024 IEEE International Conference on Big Data (BigData), Washington, DC, USA, 2024, pp. 1678-1685, doi: 10.1109/BigData62323.2024.10825143. IEEE Xplore link: https://ieeexplore.ieee.org/abstract/document/10825143}
\bibliographystyle{IEEEtran}
\bibliography{custom}

\end{document}